\documentclass{article}
\usepackage{iclr2024_conference,times}

\usepackage{amsmath,amsfonts,bm}

\def\eqref#1{equation~\ref{#1}}

\def\1{\bm{1}}

\DeclareMathAlphabet{\mathsfit}{\encodingdefault}{\sfdefault}{m}{sl}
\SetMathAlphabet{\mathsfit}{bold}{\encodingdefault}{\sfdefault}{bx}{n}

\DeclareMathOperator*{\argmin}{arg\,min}

\usepackage{hyperref}
\usepackage{url}
\usepackage{microtype}
\usepackage{graphicx}
\usepackage{subfigure}
\usepackage{booktabs}
\usepackage{array}
\usepackage{xcolor}
\usepackage{marginnote}
\usepackage{amsmath}
\usepackage{mathrsfs}
\usepackage{amssymb}
\usepackage{graphicx} 
\usepackage{caption}
\usepackage{tabularx}
\usepackage{enumitem}
\usepackage{graphicx}
\usepackage{wrapfig}
\usepackage{orcidlink}
\numberwithin{equation}{section}

\title{Towards Leveraging AutoML for Sustainable Deep Learning: A Multi-Objective HPO Approach on Deep Shift Neural Networks}

\author{Leona Hennig, Tanja Tornede, Marius Lindauer  \\
Institute of Artificial Intelligence, 
Leibniz University Hannover, Germany\\
\texttt{\{l.hennig, t.tornede, m.lindauer\}@ai.uni-hannover.de}}

\iclrfinalcopy
\begin{document}

\maketitle

\begin{abstract}
Deep Learning (DL) has advanced various fields by extracting complex patterns from large datasets.
However, the computational demands of DL models pose environmental and resource challenges. Deep shift neural networks (DSNNs) offer a solution by leveraging shift operations to reduce computational complexity at inference.
Following the insights from standard DNNs, we are interested in leveraging the full potential of DSNNs by means of AutoML techniques. 
We study the impact of hyperparameter optimization (HPO) to maximize DSNN performance while minimizing resource consumption. Since this combines multi-objective (MO) optimization with accuracy and energy consumption as potentially complementary objectives, we propose to combine state-of-the-art multi-fidelity (MF) HPO with multi-objective optimization. Experimental results demonstrate the effectiveness of our approach, resulting in models with over 80\% in accuracy and low computational cost.
Overall, our method accelerates efficient model development while enabling sustainable AI applications.
\end{abstract}

\section{Introduction}
\label{Introduction}

Deep Learning (DL) is considered one of the most promising approaches to extracting information from large data collections with complex structures. This includes performing computations in IoT environments and on edge devices \citep{DBLP:journals/network/LiOD18,DBLP:journals/pieee/ZhouCLZLZ19}. 
With the ever-increasing size and performance of such models due to the progress in science and industry, a computational cost is associated with them \citep{DBLP:journals/pieee/SzeCYE17}. 
Minimizing this cost directly affects a model's environmental impact \citep{DBLP:journals/cacm/SchwartzDSE20}. 
Thus, resources are freed up and can be used for other tasks, e.g., those with resource limitations \citep{DBLP:journals/corr/HowardZCKWWAA17}. With our approach, we contribute to DL in such low-resource settings.

Deep shift neural networks (DSNNs) offer great potential in reducing power consumption compared to traditional deep learning models \citep{DBLP:conf/cvpr/ElhoushiCSTL21}. 
Instead of floating point arithmetic, they leverage shift operations — specifically, bit shifting — as the computational unit, which boosts efficiency by replacing costly multiplication operations in convolutional networks.
We suspect that the configuration of DSNNs has a huge impact on both performance and computational efficiency.

One of the key challenges with DSNNs is determining the appropriate level of precision for the shift operations to minimize quantization errors without excessively increasing the computational load. 
In this study, we leverage DSNNs in combination with automated machine learning (AutoML) to find the optimal configuration of DSNNs in the spirit of Green AutoML \citep{DBLP:journals/jair/TornedeTHMWH23}.
This is achieved by hyperparameter optimization (HPO) using the framework SMAC3 proposed by \cite{DBLP:journals/jmlr/LindauerEFBDBRS22}, which automates the search for optimal model configurations.
Integrating multi-fidelity (MF) and multi-objective optimization (MO) techniques \citep{Belakaria2020-re} facilitates an optimal exploration of the hyperparameter space that jointly prioritizes performance and energy consumption \citep{Deb2014}. 
SMAC's MO implementation effectively balances the trade-off between achieving high predictive accuracy and minimizing energy consumption. 
The MF aspect allows for the efficient use of computational resources by evaluating configurations at varying levels of detail.
Employing tools like CodeCarbon \citep{DBLP:journals/corr/abs-1910-09700,DBLP:journals/corr/abs-1911-08354} during the training and evaluation phases provides real-time insights into the energy consumption and carbon emissions associated with each model configuration. 

Overall, we make the following contributions:
\begin{enumerate}
    \item A configuration space specific for DSNNs,
    \item a Green AutoML approach for efficiency-driven model building,
    \item insights into the design decisions for obtaining optimal tradeoffs between accuracy and energy efficiency, and
    \item combining MO and MF approaches in SMAC to improve optimization performance and computational resource utilization.
\end{enumerate}
\section{Background}
\label{background}
The following chapter introduces foundational concepts for our approach.
\subsection{Deep Shift Neural Networks}
Deep shift neural networks (DSNNs) are a novel approach to reducing the computational and energy demands of deep learning \citep{DBLP:conf/cvpr/ElhoushiCSTL21}. They achieve a considerable reduction in latency time by simplifying the network architecture such that they replace the traditional multiplication operations in neural networks by bit-wise shift operations and sign flipping, making DSNNs suitable for computing devices with limited resources.
There are two methods for training DSNNs \citep{DBLP:conf/cvpr/ElhoushiCSTL21}: DeepShift-Q (Quantization) and DeepShift-PS (Powers of two and Sign). 
DeepShift-Q involves training regular weights constrained to powers of 2 by quantizing weights to their nearest power of two during both forward and backward passes. 
DeepShift-PS directly includes the values of the shifts and sign flips as trainable parameters. 

The DeepShift-Q approach obtains the sign matrix \(S\) from the trained weight matrix \(W\) as \( S = \mathit{sign}(W) \). 
The power matrix \(P\) is the base-2 logarithm of \(W\)'s absolute values, i.e., \( P = \log_{2}(|W|) \). 
After rounding \(P\) to the nearest power of two, \( P_{\mathit{r}} = \mathit{round}(P) \), the quantized weights \(\tilde{W}_q\) are calculated by applying the sign from \(S\), shown as $\tilde{W}_q = \mathit{flip}(2^{P_{\mathit{r}}}, S)$.
The DeepShift-PS approach optimizes neural network weights by directly adapting the shift (\( \tilde{P} \)) and sign (\( \tilde{S} \)) values. 
The shift matrix \( \tilde{P} \) is obtained by rounding the base-2 logarithm of the weight values, \( \tilde{P} = \mathit{round}(P) \), and the sign flip \( \tilde{S} \) is computed as \( \tilde{S} = \mathit{sign}(\mathit{round}(S)) \). 
Weights are calculated as $\tilde{W}_{\mathit{ps}} = \mathit{flip}(2^{\tilde{P}}, \tilde{S})$, where the sign flip operation \( \tilde{S} \) assigns values of \(-1\), \(0\), or \(+1\) based on \( s \). 

\subsection{Hyperparameter Optimization}
\label{hpo}
The increasing complexity of deep learning algorithms enhances the need for automated hyperparameter optimization (HPO) to increase model performance \citep{DBLP:journals/widm/BischlBLPRCTUBBDL23}. 
Consider a dataset \( \mathcal{D} = \{(x_i,y_i)\}_{i=1}^{N} \in \mathbb{D} \subset \mathcal{X} \times \mathcal{Y} \), where \( \mathcal{X} \) is the instance space and \( \mathcal{Y} \) is the target space, and a hyperparameter configuration space \( \Lambda = \{\lambda_1, \ldots, \lambda_L\} \), \( L \in \mathbb{N} \). In our study, \( \mathcal{M} \) denotes the space of possible DSNN models. An algorithm \( \mathcal{A} : \mathbb{D} \times \Lambda \rightarrow \mathcal{M} \) trains a model \( M \in \mathcal{M} \), instantiated with a configuration of \( L \) hyperparameters sampled from \( \Lambda \), on a training subset of \( \mathcal{D} \). The dataset \( \mathcal{D} \) is split into training, validation, and test sets: \( \mathcal{D}_{\textit{train}}, \mathcal{D}_{\textit{val}}, \) and \( \mathcal{D}_{\textit{test}} \) respectively. The performance of the algorithm is assessed via an expensive-to-evaluate loss function \( \mathcal{L} : \mathcal{M} \times \mathbb{D} \rightarrow \mathbb{R} \). The optimization objective of HPO is to find the configuration \( \lambda^* \in \Lambda \) with minimal validation loss \( \mathcal{L} \), such that:
\begin{equation}
\lambda^* \in \argmin_{\lambda \in \Lambda} \mathcal{L}\big(\mathcal{A}(\mathcal{D}_{\textit{train}},\lambda), \mathcal{D}_{\textit{val}}\big).
\end{equation}
This process tunes hyperparameters based on validation set performance. 
Models are trained on \(\mathcal{D}_{\textit{train}}\) and optimized using \(\mathcal{D}_{\textit{val}}\). Finally, the model's performance is assessed on \(\mathcal{D}_{\textit{test}}\).

\subsection{Bayesian Optimization}
\label{bo}

Bayesian Optimization (BO) is a strategy for global optimization of black-box loss functions $\mathcal{L}:\mathcal{M}\times\mathbb{D}\xrightarrow{}\mathbb{R}$ that are expensive to evaluate \citep{DBLP:journals/jgo/JonesSW98}.
BO uses a surrogate model $\mathcal{S}$, a probabilistic model to approximate the loss function, commonly given by a Gaussian Process or a Random Forest \citep{DBLP:books/lib/RasmussenW06,DBLP:conf/lion/HutterHL11}. 
An acquisition function $\mathcal{C}:\Lambda\xrightarrow{}\mathbb{R}$ guides the search for the next optimal evaluation points by balancing the exploration-exploitation trade-off, based on the set of previously queried points $\{(\lambda_1,\mathcal{L}_1),...,(\lambda_{m-1},\mathcal{L}_{m-1})\}$ at time $m$.
$\mathcal{L}$ is only evaluated at certain points, with which the surrogate model is updated. 

\subsection{Multi-Fidelity Optimization}
\label{multi-fidelity}

Since it is not feasible to fully train multiple configurations of DSNNs for comparison due to computational efficiency, we employ a multi-fidelity (MF) approach \citep{DBLP:journals/jmlr/LiJDRT17}, which is a common strategy in AutoML to navigate the trade-off between performance and approximation error \citep{DBLP:books/sp/HKV2019}. 
MF approaches train cheap-to-evaluate proxies of black-box functions following different heuristics, e.g., allocating a small number of epochs to many configurations in the beginning and training the best-performing ones on an increasing number of epochs. 
Formally, we define a space of fidelities $\mathcal{F}$ and aim to minimize a high-fidelity function $F\in\mathcal{F}$ \citep{DBLP:journals/jair/KandasamyDOSP19}:
\begin{equation}
    \min_{\lambda\in\Lambda} F(\lambda)\, .
\end{equation}
We aim to approximate $F\in\mathcal{F}$, using a series of lower-fidelity, less expensive approximations 
$\{f(\lambda)_1,\ldots,f(\lambda)_j\}$, where $j$ denotes the total number of fidelity levels.
The allocated resources for evaluating a model's performance at various fidelities are referred to as a budget. 
MF assumes that the highest fidelities approximate the black-box function best.

\subsection{Multi-Objective Optimization}
Multi-Objective Optimization (MO) addresses problems involving multiple, often competing objectives. This approach is used in scenarios where trade-offs between two or more conflicting objectives must be navigated, such as, in the context of DSNNs, enhancing accuracy alongside reducing energy consumption. MO aims to identify Pareto-optimal solutions \citep{Deb2014}. To ensure effective approximation of the Pareto front, new points are added based on the current observation dataset \( \mathcal{D}_\mathit{obs} = \{(\lambda_1,\mathcal{L}(\lambda_1)), \ldots, (\lambda_n,\mathcal{L}(\lambda_n))\} \) at time $n$. These points augment the surface formed by the non-dominated solution set \( D^\star_n \), which satisfies the condition for $d$ objective variables and a loss function $\mathcal{L} = (\mathcal{L}_1,\ldots,\mathcal{L}_d)$ \citep{DBLP:journals/corr/abs-1905-02370}: For all $\lambda, (\lambda,\mathcal{L}(\lambda)) \in \mathcal{D}^\star_n \subset \mathcal{D}_n$ and $\lambda', (\lambda',\mathcal{L}(\lambda')) \in \mathcal{D}_n$ exists $k \in \{1,\ldots,d\}$, such that $\mathcal{L}_k(\lambda) \leq \mathcal{L}_k(\lambda')$.

\section{Approach}
\label{approach}
To computationally enhance Deep shift neural networks (DSNN) via AutoML, we employ multi-fidelity optimization (MF).
We introduce different fidelities to the training process by increasing the number of shift layers in the model. 
The initially trained models will include fewer shift layers, which will increase during the optimization process. We assume this will guide the search toward the highest-performing shift networks since a low number of shift layers introduces the least numerical inaccuracy and mathematical uncertainty to the model. We aim to investigate whether picking the best-performing models under low model-intrinsic uncertainty continues to show improved performance when combined with more shift layers.

In Section \ref{background}, we explained the MF approach. One algorithm that formalizes its heuristics is successive halving \citep{jamieson-aistats16a}, where $n_c$ configurations are trained on an initial small budget $b_I$. 
After that, the $\nu/(\nu+1)$ best-performing configurations are trained on a budget of $\nu b_I/(\nu +1)$ until the best configuration is determined. 
We address the trade-off between $b_I$ and $n_c$, or between approximation error and exploration inherent in successive halving, using the HyperBand algorithm for MF.
HyperBand \citep{DBLP:journals/jmlr/LiJDRT17} runs successive halving in multiple brackets, where each bracket provides a combination of $n_c$ and a fraction of the total budget per configuration so that they sum up to the total budget. 

We extend this to multi-fidelity multi-objective optimization (MFMO). We simultaneously address the accuracy of the model as well as its energy consumption.
The goal is to optimize the performance of the DSNNs, ensuring they achieve high accuracy and robustness in their predictive capabilities; secondly, to minimize energy consumption, a critical factor given the environmental implications of computational efficiency.
To this end, we implement a two-dimensional objective function $f_{MO}:\Lambda\xrightarrow{}\mathbb{R}^2,\quad f_{M0}(\lambda) = \big(f_{loss}(\lambda), f_{emission}(\lambda)\big)$, where, given a configuration $\lambda\in\Lambda$, $f_{loss}(\lambda)$ aims to minimize the loss, enhancing the model's accuracy, and $f_{emission}(\lambda)$ seeks to minimize the energy consumption during training and inference, promoting environmental sustainability.
We aim to solve the following optimization problem:

\begin{equation}
    \argmin_{\lambda\in\Lambda} f_{M0}(\lambda)\, .
\end{equation}

SMAC3 uses a mean aggregation strategy by calculating the objective's arithmetic mean to aggregate multiple objectives into a single scalar value for MF optimization.
We use the ParEGO algorithm \citep{DBLP:journals/tec/Knowles06} to compute Pareto optimal objectives. It transforms the multi-objective problem into a series of single-objective problems by introducing varying weight parameters for the objectives in each iteration of HyperBand. Thus optimizing a different scalarization per evaluation to approximate the Pareto front.
The resulting single-objective optimization function can now be evaluated in an MF setting.

\section{Experiments}
\label{experiments}

In the following section, we detail the setup and methodology used to evaluate our approach discussed in Section \ref{approach}, focusing on optimizing deep shift neural networks (DSNNs) through multi-fidelity, multi-objective optimization (MFMO) as well as multi-fidelity optimization (MF). We discuss how our approach successfully navigates the model performance and environmental impact trade-offs.

\subsection{Evaluation Setup}
\label{eval setup}
We train and evaluate our models on the Cifar10 dataset \citep{krizhevsky2009learning}, using NVIDIA A100 GPUs.
For hyperparameter optimization (HPO), we rely on using SMAC3~\citep{DBLP:journals/jmlr/LindauerEFBDBRS22}.
To incorporate the environmental impact into our HPO workflow, we use the CodeCarbon emissions tracker (\cite{DBLP:journals/corr/abs-1910-09700}, \cite{DBLP:journals/corr/abs-1911-08354}) to track carbon emissions from computational processes by monitoring energy use and regional energy mix in $g\mathit{CO}_{2}\mathit{eq}$, grams of $\mathit{CO}_2$ equivalent \citep{DBLP:journals/jair/TornedeTHMWH23}.
We chose a pre-trained ResNet20 \citep{DBLP:conf/cvpr/HeZRS16}.

\subsection{Results}
\label{results}
\begin{wrapfigure}{r}{0.5\textwidth}
  \begin{center}
    \includegraphics[width=0.5\textwidth]{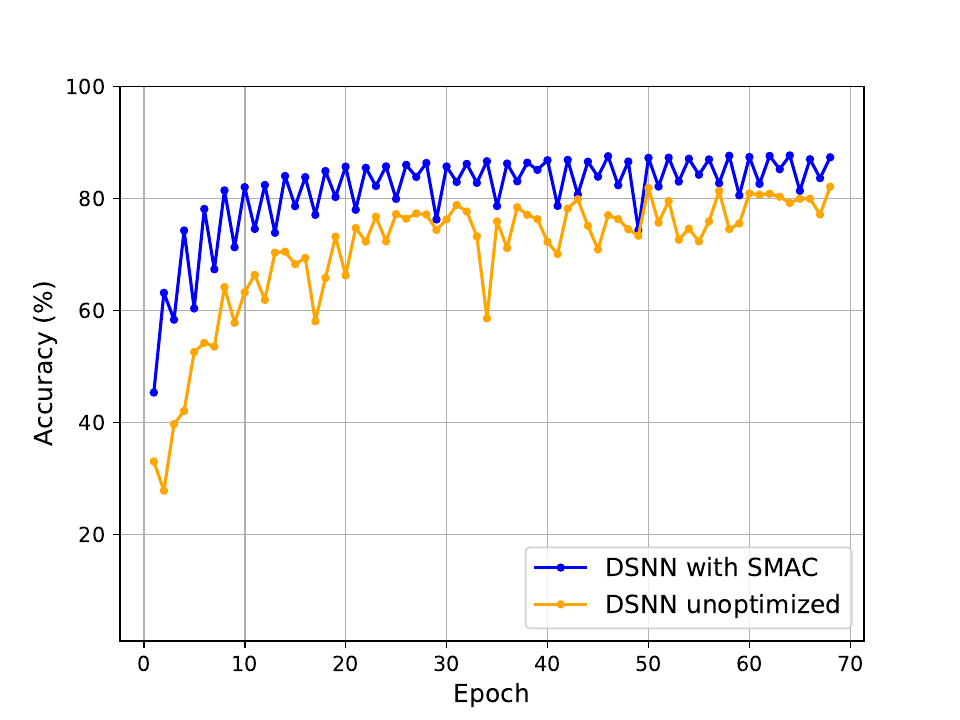}
  \end{center}
  \captionsetup{justification=centering}
  \caption{Test accuracy of the default DSNN and a configured DSNN}
  \label{fig:results}
\end{wrapfigure}
Figure \ref{fig:results} displays the Top-1 test accuracy of a DSNN trained with the default configuration from Table \ref{table:model_config} as defined by \cite{DBLP:conf/cvpr/ElhoushiCSTL21}, as well as the test accuracy from the configured DSNN according to SMAC3 with MF using HyperBand, with the number of shift layers inducing the fidelities. The Pareto-optimal Solution No. 1 configuration from Table \ref{table:model_config} is used. The MF algorithm, trained on the same seed as MFMO, identifies it as the optimal model configuration out of fifty evaluated configurations. While volatile to a certain degree, the plot clearly shows that the model's performance with the optimal configuration surpasses that of the model instantiated with the default configuration by at least three percent. This verifies our approach that MF can successfully be executed with DSNN-specific budgets such as the shift depth.

Furthermore, it confirms our assumption from Section \ref{approach} that the models can be evaluated under low model-intrinsic uncertainty and continue to perform well when instantiated with more shift layers.

Table \ref{table:model_config} shows two Pareto optimal solutions out of 33 evaluated configurations with our MFMO approach.
Evaluating solution No. 1 results in a Top-1 accuracy of 83.50\% and 0.1661 $g\mathit{CO}_{2}\mathit{eq}$. Solution No. 2 yields a Top-1 accuracy of 84,67\%, producing 0.1673 $g\mathit{CO}_{2}\mathit{eq}$.
Note that they were trained for fewer epochs due to computational limitations. Both configurations yield good performance results while keeping emissions low.

The DSNN-specific model hyperparameters include activation integer bits, activation fraction bits, weight bits, and shift depth.
Solution No.1 includes fewer shift layers with fewer bits representing the weights activation function values.
Activation bits and weight bits seem to affect emissions and model performance similarly.
Fewer shift layers increase the computational demand of a model.
However, the reduced number of bits used to replace floating-point operations (FLOPs) seems to compensate for this since solution No. 1 produces approximately the same amount of emissions as solution No. 2.
This one contains more than three times the number of shift layers, using only four convolutional layers with FLOPs, but uses substantially more representation bits.
There seems to be a trade-off between the number of shift layers and the number of representation bits that, when navigated efficiently, yields model configurations with satisfying performance and energy consumption.
This supports our claim that our MFMO approach is promising for building high-performance, energy-efficient DSNNs.
\begin{table}
\centering
\footnotesize 
\setlength{\tabcolsep}{3pt} 
\renewcommand{\arraystretch}{0.8} 
\begin{tabular}{@{}>{\raggedright\arraybackslash}p{3.5cm} >{\raggedright\arraybackslash}p{4cm} >{\raggedright\arraybackslash}p{1.5cm}>{\raggedright\arraybackslash}p{1.5cm} >{\raggedright\arraybackslash}p{1.5cm}@{}}
\toprule
\textbf{Parameter} & \textbf{Search Space} & \textbf{Default} & \textbf{POS 1} & \textbf{POS 2} \\
\midrule
Batch Size & [32, 128] & 128 & 97 & 102 \\
Optimizer & SGD, Adam, Adagrad, Adadelta, RMSProp, RAdam, Ranger & SGD & Ranger & SGD \\
Learning Rate & [0.001, 0.1] & 0.1 & 0.0847 & 0.0798 \\
Momentum & [0.0, 0.9] & 0.9& 0.5016 & 0.6738 \\
Epochs & [5, 100] & 200 & 67 & 91 \\
Weight Bits & [2, 8] & 5 & \textbf{4} & \textbf{7} \\
Activation Integer Bits & [2, 32] & 16 & \textbf{4} & \textbf{29} \\
Activation Fraction Bits & [2, 32] & 16 &\textbf{4} & \textbf{10} \\
Shift Depth & [0, 20] & 20 & \textbf{5} & \textbf{16} \\
Shift Type & Q, PS & PS & Q & Q \\
Rounding & deterministic, stochastic & deterministic & stochastic & deterministic \\
Weight Decay & [1e-6, 1e-2] & 0.0001 & 0.0026 & 0.00012 \\
Accuracy (in \%) & &  & 83.50 & 84.67 \\
Emissions (in $g\mathit{CO}_{2}\mathit{eq}$) & &  & 0.1661 & 0.1673 \\
\bottomrule
\end{tabular}
\caption{MFMO findings, incl. two Pareto optimal solutions (POS 1 \& 2)}\label{table:model_config}
\end{table}

\section{Conclusion}
\label{conclusion}

In this work, we present our Green AutoML approach towards the sustainable optimization of DSNNs through a multi-fidelity, multi-objective (MFMO)  HPO framework.
We address the critical intersection between advancing the capabilities of Deep Learning and environmental sustainability. By leveraging AutoML tools and integrating the environmental impact as an objective, we navigate the trade-off between model performance and efficient resource utilization.

Our experimental results highlight the potential of our approach. We successfully optimized a DSNN to achieve high accuracy while minimizing energy consumption.
We have delivered a thorough configuration space for DSNNs, introduced a Green AutoML approach for efficiency-driven model development, and provided valuable insights into the design decisions. Future work includes extending our approach to multiple benchmarks and neural network architectures to sufficiently verify our approach on a broad spectrum of model designs and applications. This way, we hope to gain more insight into critical design decisions for DSNNs and other models built for energy-efficient computations. Revisiting our MFMO implementation to find a more efficient way for ParEGO and Hyperband to intertwine, e.g. by finding a more effective way to assign budgets and weights to configurations, is subject to further work to lighten the computational load when computing the MFMO Pareto fronts. Furthermore, we will investigate more DSNN-specific fidelity types and MO algorithms to achieve further reductions in model emissions.

\subsubsection*{Acknowledgments}
This work was supported by the German Federal Ministry of the Environment, Nature Conservation, Nuclear Safety and Consumer Protection (GreenAutoML4FAS project no. 67KI32007A).

\bibliography{iclr2024_conference,strings,lib,proc}
\bibliographystyle{iclr2024_conference}

\end{document}